\documentclass{article}

\PassOptionsToPackage{numbers, compress}{natbib}
\usepackage[final,nonatbib]{neurips_2024}

\usepackage[utf8]{inputenc} 
\usepackage[T1]{fontenc}    
\usepackage{hyperref}       
\usepackage{url}            
\usepackage{booktabs}       
\usepackage{amsfonts}       
\usepackage{nicefrac}       
\usepackage{microtype}      
\usepackage{xcolor}         
\usepackage{xspace}
\usepackage{graphicx}
\usepackage{float}
\usepackage{booktabs} 
\usepackage{multirow} 
\usepackage{utfsym}
\usepackage{arydshln}
\usepackage[normalem]{ulem}
\usepackage{amsmath}
\usepackage{amssymb}
\def\swabbr{CSKV\xspace}

\title{\swabbr: Training-Efficient Channel Shrinking for KV Cache in Long-Context Scenarios}

\author{%
  Luning Wang$^{1,2}$, Shiyao Li$^{1,2}$, Xuefei Ning$^{1}$, Zhihang Yuan$^{2}$,\\ \textbf{Shengen Yan$^{2}$, Guohao Dai$^{3,2}$, Yu Wang$^{1}$} \\ \\
  $^{1}$Tsinghua University \\
  $^{2}$Infinigence-AI \\
  $^{3}$Shanghai Jiao Tong University
}

\begin{document}

\maketitle

\begin{abstract}
\label{sec:abs}
Large Language Models (LLMs) have been widely adopted to process long-context tasks.
However, the large memory overhead of the key-value (KV) cache poses significant challenges in long-context scenarios.
Existing training-free KV cache compression methods typically focus on quantization and token pruning, which have compression limits, and excessive sparsity can lead to severe performance degradation.
Other methods design new architectures with le
ss KV overhead but require significant training overhead.
To address the above two drawbacks, we further explore the redundancy in the channel dimension and apply an architecture-level design with minor training costs.
Therefore, we introduce \textbf{\swabbr}, a training-efficient \textbf{\underline{C}}hannel \textbf{\underline{S}}hrinking technique for \textbf{\underline{KV}} cache compression:
(1) We first analyze the singular value distribution of the KV cache, revealing significant redundancy and compression potential along the channel dimension.
Based on this observation, we propose using low-rank decomposition for key and value layers and storing the low-dimension features.
(2) To preserve model performance, we introduce a bi-branch KV cache, including a window-based full-precision KV cache and a low-precision compressed KV cache.
(3) To reduce the training costs, we minimize the layer-wise reconstruction loss for the compressed KV cache instead of retraining the entire LLMs.
Extensive experiments show that \swabbr can reduce the memory overhead of the KV cache by 80\% while maintaining the model's long-context capability.
Moreover, we show that our method can be seamlessly combined with quantization to further reduce the memory overhead, achieving a compression ratio of up to 95\%. Code is available at \url{https://github.com/wln20/CSKV}.
\end{abstract}

\section{Introduction}
\label{sec:intro}
Large Language Models (LLMs) have been widely adopted in various natural language processing tasks, particularly those requiring long-context capabilities, such as document analysis and fact retrieval~\cite{longcontextsurvey}. However, the key-value (KV) cache mechanism used in transformer-based LLMs poses significant efficiency challenges as its memory overhead grows linearly with the sequence length, often replacing the weights to be the memory bottleneck in long-context scenarios. For instance, processing a sequence with 200K tokens using LLaMA-2-7B~\cite{llama2} results in a KV cache occupying around 100GB, compared to 14GB required for model weights. Compressing the KV cache by over 10× is necessary to fit such a sequence on a single NVIDIA RTX 4090 GPU with 24GB of memory.

Existing KV cache compression methods, mainly training-free techniques like token pruning~\cite{h2o,scissorhands,streamingllm,kim2022learned} and quantization~\cite{evaluatingquant,kivi,lin2024qserve,sheng2023flexgen}, struggle to maintain model performance at high compression ratios, particularly in long-context tasks. Alternatively, training-required techniques, such as MLA~\cite{mla} and cache sharing~\cite{yoco,cla}, offer higher compression ratios but at the cost of significant retraining and are typically unable to be integrated with existing pre-trained models.

Inspired by MLA, we observe significant redundancy in the large channel dimensions of the KV cache, evidenced by the long-tailed distribution of singular values in the key and value caches (Details in Appendix). Experiments reveal that removing the smallest 50\% of these singular values results in less than 1\% average accuracy loss on the MMLU~\cite{mmlu} benchmark (from 0.458 to 0.449). 

Given this redundancy, we propose \textbf{\swabbr}, a training-efficient \textbf{\underline{C}}hannel \textbf{\underline{S}}hrinking technique for the \textbf{\underline{KV}} cache, designed to balances high compression ratios with low training costs. To sum up, we have the following contributions:

\begin{itemize}
    \item To reduce the memory overhead of the KV cache while maintaining the performance, we design a \textbf{bi-branch KV cache} by preserving the recently used KV cache with original dimensions and reducing the dimension of the historical KV cache. 
    \item To further improve the performance without significant training overhead, we propose an effective \textbf{SVD-based initialization} technique and train LLMs in a layer-wise manner by minimizing the reconstruction loss. 
    \item Extensive experimental results demonstrate that our method can achieve an 80\% KV cache compression ratio while maintaining the model's long-context capability. We further demonstrate that our method can be seamlessly combined with 4-bit quantization, showcasing its power in achieving a total compression ratio of 95\%.
\end{itemize}

\section{Method}
\label{sec:method}
\subsection{Inference with Bi-Branch KV Cache}
\textbf{To reduce the memory overhead}, we design to reduce the memory overhead of the KV cache by using low-rank decomposition for both the Key and Value weight matrix. 
Without loss of generality, we will detail the workflow of compressing the key cache, as the process is identical to that of the value cache.

\begin{figure*}[!t]
    \centering    
    \includegraphics[width=0.8\textwidth, height=3.5cm]{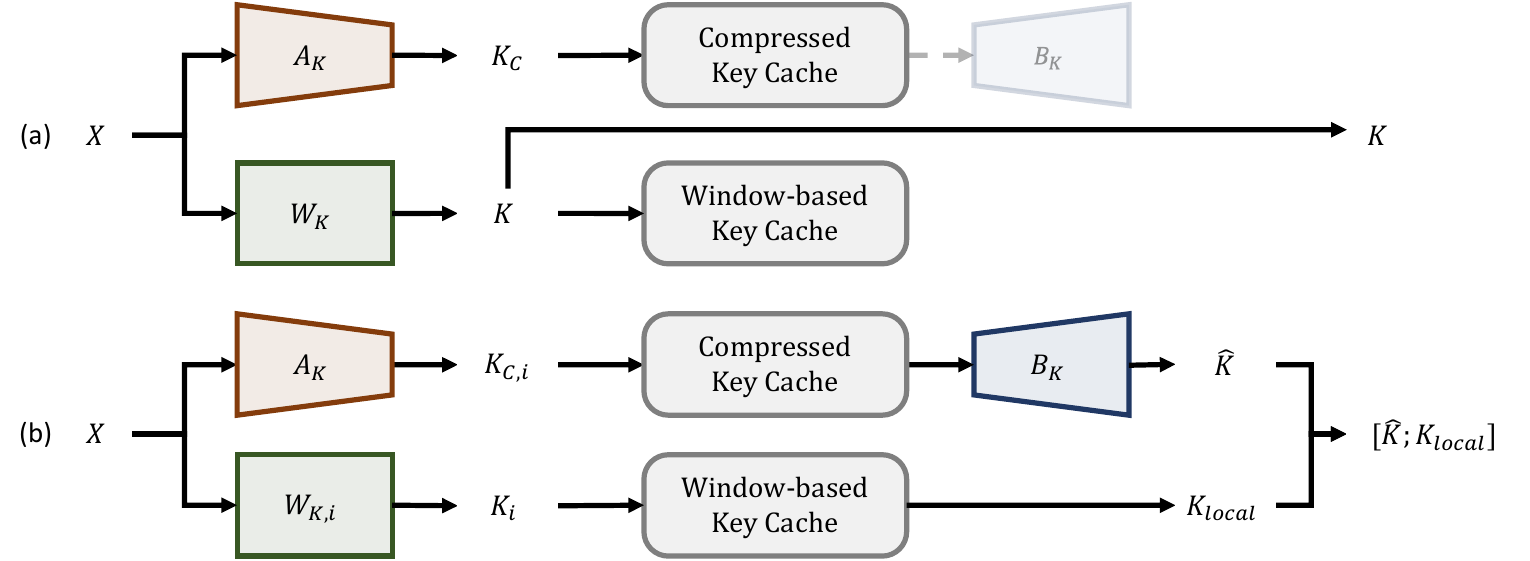}
    \caption{The overview of the inference process. (a) The prefilling stage. (b) The decoding stage.}
    \label{fig:prefill_decode}
\end{figure*}

As shown in Figure~\ref{fig:prefill_decode}, we use two matrices, $A_K \in R^{h_{in} \times h_{comp}}$ and $B_K \in R^{h_{comp} \times h_{out}}$, to approximate the weight matrix of $W_K \in R^{h_{in} \times h_{out}}$. 
Here the $h_{in},h_{out},h_{comp}$ are the input dimension of $W_K$, the output dimension of $W_K$, and the intermediate dimension of the low-rank decompression.
Keeping the $h_{comp}$ smaller than $h_{out}$ and \textbf{storing the intermediate features as the compressed Key cache}, we can significantly save the memory overhead, especially in the long context scenario.

\textbf{To maintain the high performance}, we propose to follow the prior research by preserving the recently used tokens~\cite{window_based,streamingllm} because they are crucial for accurate next-token prediction.
To prevent the degradation of this local information during inference, we propose \textbf{the bi-branch KV cache} that preserves the recently used tokens effectively during both the prefilling and decoding stages.
With a pre-defined window size $l_w$, we compress the KV cache only after the tokens fill a complete window while retaining the residual tokens in their original hidden dimensions.

Specifically, for the prefilling stage, as shown in Figure~\ref{fig:prefill_decode}(a), given an input sequence with $n$ tokens, we first use the $A_K$ to generate the compressed Key matrix and store it in the Compressed Key Cache $K_C$.
In this case, the Compressed Key Cache contains all of the historical information of the given sentence.
On the other branch, we use the original $W_K$ to generate the full-precision Key matrix $K$ for computation, which can guarantee that the computation results of the prefilling stage are the same as the original LLMs.
Then, we only store the full-precision Key activation of the last $m$ tokens $K_{local}$ to preserve the local information for the decoding stage.

Moreover, during the decoding stage, as shown in Figure~\ref{fig:prefill_decode}(b), we only process one token during each forward pass. 
We take the process of the $(n+1)$-th token as an example.
For the cache update, we compute both the compressed Key activation $K_{C}$ and full-precision Key activation $K$ and update both Key caches with the new activations.
In this case, the compressed Key cache has $(n+1)$ tokens, and the full-precision Key cache has $(m+1)$ tokens.
To get the $(n+1)$ tokens' Key matrix, we use the $(m+1)$ tokens from the full-precision Key cache as $K_{local}$ and use the $B_K$ to process the oldest $(n-m)$ tokens in the compressed Key cache as $\hat{K}$.
By concatenating the $\hat{K}$ and $K_{local}$, we can get the target Key matrix for attention computation.
Finally, we remove the oldest token from the full-precision Key cache to keep the window size as $m$.

\subsection{Efficient Fine-tuning by SVD-based Initialization}

\begin{figure*}[!t]
    \centering    
    \includegraphics[width=0.75\textwidth, height=2.0cm]{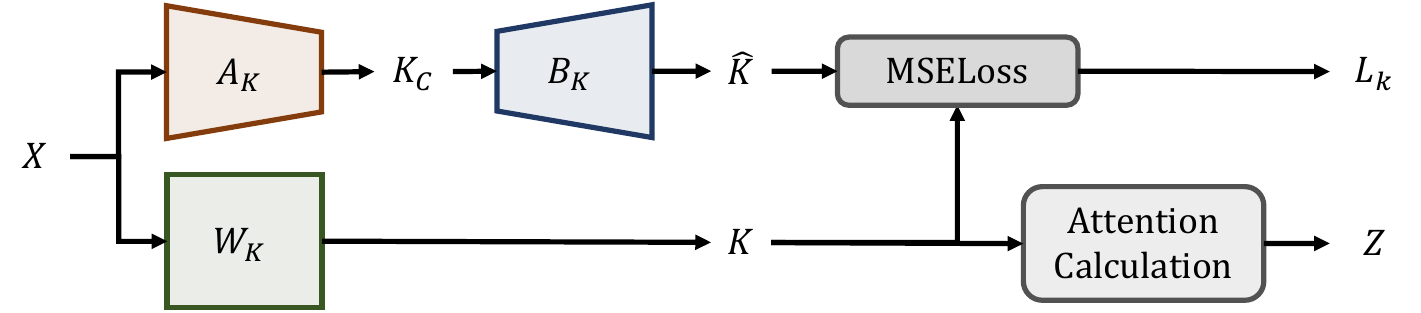}
    \caption{The overview of the efficient layer-wise reconstruction fine-tuning.}
    \label{fig:finetune}
\end{figure*}

Directly applying low-rank decomposed weight matrices for KV cache compression would result in the degradation of model performance when the compression ratio becomes high.
To further enhance the model performance, we propose to introduce an efficient training process. 
We find that the initialization method to the proposed $A_K$ and $B_K$ is of great importance for convergence and final performance.
In this case, we proposed to use the ASVD-based decomposition results for initialization.
As shown in Figure~\ref{fig:finetune}, we train LLMs in a layer-wise manner by minimizing the layer-wise reconstruction loss for the compressed keys and values.

Specifically, for each layer, we can use the $W_K$ to generate the full-precision Key matrix $K=XW_K$ and use $A_K$, $B_K$ to generate the lossy key matrix $\hat{K} = X A_K B_K$.
The local reconstruction loss of this layer could be defined as Equation~\ref{eq:layer_loss}:
\begin{equation}
    \begin{aligned}
        L_K = \mathrm{MSELoss}(K,\hat{K})
    \end{aligned}
    \label{eq:layer_loss}
\end{equation}
where $L_K$ denotes the loss of keys in this layer, and $\mathrm{MSELoss}(\cdot, \cdot)$ is the Mean Square Error (MSE) loss function.
Finally, define the loss of keys and values in the $i$-th layer as $L_{K,i},L_{V,i}$, the loss for the whole model is shown in Equantion~\ref{eq:all_loss}:
\begin{equation}
    \begin{aligned}
        \mathcal{L}_{all}=\sum_{j=0}^{n_l}\left(L_{K,j}+L_{V,j}\right)
    \end{aligned}
    \label{eq:all_loss}
\end{equation}
where $\mathcal{L}_{all}$ denotes the loss for the whole model, and $n_l$ denotes the number of layers.

\section{Experiment}
\label{sec:exp}
\subsection{Experimental Setup}
\label{sec:exp-setup}
We evaluate our method on LongChat-7B-v1.5-32k~\cite{longchat} and Mistral-7B-Instruct-v0.2~\cite{mistral}. 
We evaluate our method on three widely-used long-context benchmarks: LongEval~\cite{longchat}, LongBench~\cite{longbench}, and LVEval~\cite{lveval}.
For comparison, we include results from StreamingLLM~\cite{streamingllm}, H$_2$O~\cite{h2o}~\footnote{Here we only compare the effect of H$_2$O on Longchat-7b-v1.5-32k, as it only supports LLaMA architecture in its official implementation.}, and ASVD~\cite{asvd}.
The first two are token pruning methods, while the latter is a SOTA channel-shrinking method.
More details can be found in the Appendix.

\subsection{Main Results}
\label{sec:eval}
We apply compression ratios of 50\% and 80\% consistently for both keys and values. The results are presented in Table~\ref{tab:comprehensive}.

According to the evaluation results in table~\ref{tab:comprehensive}, the token pruning methods are especially not skilled in retrieval tasks like LongEval, even at a 50\% compression ratio, when ASVD and CSKV only incur minor performance loss. As the compression ratio reaches 80\%, all methods except for CSKV suffer great performance degradation on all three tasks. To dive deeper, we examine the failure cases of token pruning methods, and found that although the model could generate coherent sentences based on instructions, a great deal of the retrieved answers deviate from the ground truth by a small portion, like answering "4244" when the label is "42440", or give an irrelevant answer such as "1386". This might be caused by their token eviction mechanisms which inherently have to discard the information of some tokens completely, facing great risk of losing the ground truth information. In contrast, the abundant failure cases of ASVD at 80\% compression are mainly caused by the loss of the model's language modeling capabilities, like responding with dozens of tokens that could hardly form a sentence. Different from the aforementioned methods, CSKV consistently enables the model to generate instruction-following responses and give accurate answers on either retrieval tasks or QA tasks, showing its superior capability of keeping the model's long-context abilities even at high compression ratios.

\subsection{Ablation Studies}
\label{sec:ablation}
We conduct several ablation studies to further explore the potential of our method, and the main conclusions include: 1) The SVD-based initialization methods is crucial to the success of training; 2) The model performance is positively correlated with the window size, while the benefit would become less significant after it reaches a certain level; 3) In most cases, it would be better to compress the key cache more than the value cache given a certain budget; 4) \swabbr could be seamlessly integrated with 4-bit QAT with very small performance loss. See Appendix for details.

\begin{table}[!t]
    \centering
    \caption{Performance of models with CSKV on long-context benchmarks.}
    \resizebox{\textwidth}{!}{
    \begin{tabular}{ccccccccccc}
         \toprule
         \multirow{2}{*}{Model} & \multirow{2}{*}{C. Ratio} & \multirow{2}{*}{Method} & \multicolumn{4}{c}{LongEval $\uparrow$} & \multicolumn{3}{c}{LongBench $\uparrow$} & \multicolumn{1}{c}{LV-Eval $\uparrow$} \\
         \cmidrule(lr){4-7}\cmidrule(lr){8-10}\cmidrule(lr){11-11}
         & & & 4k & 6k & 8k & 10k & 0-4k & 4-8k & 8k+ & 16k \\
         \midrule
         \multirow{9}{*}{Longchat-7b-v1.5-32k} & 0\% & - & 1.00 & 1.00 & 0.98 & 0.98 & 0.46 & 0.43 & 0.46 & 0.13 \\
         \cmidrule(lr){2-11}
         & \multirow{4}{*}{50\%} & StreamingLLM & 0.12 & 0.16 & 0.06 & 0.20 & 0.37 & 0.39 & 0.40 & 0.09 \\
         & & H$_2$O & 0.62 & 0.56 & 0.52 & 0.50 & 0.40 & 0.38 & 0.38 & 0.09 \\
         & & ASVD & 0.92 & \textbf{0.96} & 0.92 & \textbf{0.94} & 0.44 & 0.41 & 0.43 & 0.11 \\
         & & \textbf{CSKV (Ours)} & \textbf{0.98} & 0.94 & \textbf{0.96} & \textbf{0.94} & \textbf{0.46} & \textbf{0.42} & \textbf{0.45} & \textbf{0.12} \\
         \cmidrule(lr){2-11}
         & \multirow{4}{*}{80\%} & StreamingLLM & 0.06 & 0.06 & 0.02 & 0.02 & 0.31 & 0.35 & 0.39 & 0.06 \\
         & & H$_2$O & 0.18 & 0.24 & 0.26 & 0.10 & 0.34 & 0.30 & 0.32 & 0.05 \\
         & & ASVD & 0.26 & 0.12 & 0.06 & 0.04 & 0.36 & 0.31 & 0.32 & 0.04 \\         
         & & \textbf{CSKV (Ours)} & \textbf{0.92} & \textbf{0.94} & \textbf{0.94} & \textbf{0.90} & \textbf{0.43} & \textbf{0.40} & \textbf{0.41} & \textbf{0.10} \\
         \cmidrule(lr){1-11}         
         \multirow{7}{*}{Mistral-7b-instruct-v0.2} & 0\% & - & 1.00 & 1.00 & 0.98 & 0.94 & 0.50 & 0.47 & 0.45 & 0.20 \\
         \cmidrule(lr){2-11}
         & \multirow{3}{*}{50\%} & StreamingLLM & 0.06 & 0.12 & 0.04 & 0.14 & 0.39 & 0.38 & 0.37 & 0.12 \\
         & & ASVD & \textbf{1.00} & 0.98 & 0.92 & \textbf{0.94} & 0.49 & 0.45 & 0.44 & 0.17 \\
         & & \textbf{CSKV (Ours)} & \textbf{1.00} & \textbf{1.00} & \textbf{0.96} & \textbf{0.94} & \textbf{0.50} & \textbf{0.47} & \textbf{0.47} & \textbf{0.20} \\
         \cmidrule(lr){2-11}
         & \multirow{3}{*}{80\%} & StreamingLLM & 0.06 & 0.04 & 0.00 & 0.04 & 0.34 & 0.34 & 0.33 & 0.06 \\
         & & ASVD & 0.04 & 0.00 & 0.04 & 0.00 & 0.33 & 0.29 & 0.29 & 0.05 \\
         & & \textbf{CSKV (Ours)} & \textbf{0.98} & \textbf{0.96} & \textbf{0.90} & \textbf{0.92} & \textbf{0.45} & \textbf{0.42} & \textbf{0.41} & \textbf{0.17} \\
        \bottomrule
    \end{tabular}}
    \label{tab:comprehensive}
\end{table}

\section{Limitation and Future Directions}
\label{sec:limit}
While demonstrating competitive performance, the proposed method's compression ratio assignment is user-defined and might not be optimal, offering the potential to achieve higher compression ratios. Future work could explore the application of automated search algorithms to dynamically assign compression ratios to individual layers, accounting for their varying sensitivity to compression. Similarly, automated strategies could optimize memory budget allocation for keys and values, maximizing performance within a given constraint. We leave those directions for future works to explore.

\clearpage

\clearpage
\appendix

\section*{Appendix}
\subsection*{A. Distribution of Singular Values of key cache}
\label{sec:dist_kv}
We visualize the distribution of singular values of key cache in the 14-th layer of LLaMA-2-7B-chat model, using data randomly sampled from the Pile~\cite{pile} dataset. 
We find that the singular value of the key cache has a significant long-tailed distribution, and a similar phenomenon also appears in the value cache.
In this case, only a tiny fraction of singular values have large magnitudes, while the vast majority are around zero, which can be removed without significant degradation of model performance. 

\begin{figure}[H]
    \centering    
    \includegraphics[width=8.0cm, height=8.0cm]{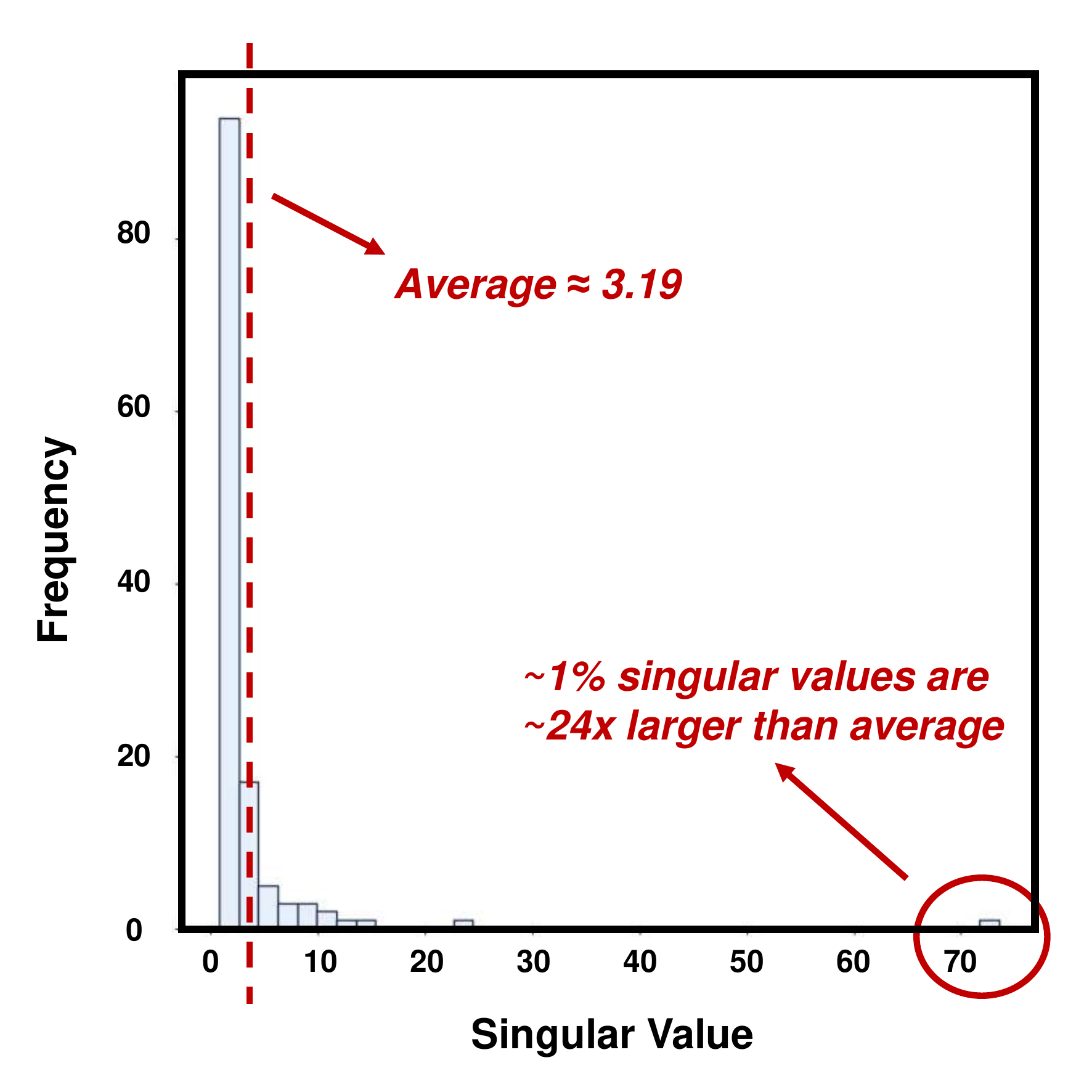}
    \caption{Distribution of of Singular Values of key cache.}
    \label{fig:kvcache_svd_all}
\end{figure}

\subsection*{B. Details of Experimental Setup}
\label{app:exp_setup_detail}
We evaluate our method on widely used long-context models, including LongChat-7B-v1.5-32k~\cite{longchat} and Mistral-7B-Instruct-v0.2~\cite{mistral}. For fine-tuning, we use a scaled-down version of the Pile~\cite{pile} dataset~\cite{Small-the-pile} and is conducted with both the epoch and batch size set to 1, using the AdamW optimizer with an initial learning rate of 5e-5. The entire fine-tuning process for each 7B model is completed within 90 minutes on a single NVIDIA A100-80G GPU, resulting in minimal training costs.
We initialize the model with ASVD~\cite{asvd}, selecting 256 samples from the fine-tuning dataset as calibration data. We set $\alpha=0.5$ and use the Absolute Mean Value method for configuring the scaling matrix $S$.

The evaluation of our method is performed on three widely-used long-context benchmarks, including LongEval~\cite{longchat}, LongBench~\cite{longbench} and LVEval~\cite{lveval}. Specifically, we choose the 200,300,400,500 lines subsets in LongEval (with an average length of 4k,6k,8k,10k), the qasper, hotpotqa, multifieldqa\_en, gov\_report, triviaqa subset of LongBench-E, along with the 16K subset of LVEval. To compare the results with other methods, we choose StreamingLLM\cite{streamingllm}, H$_2$O\cite{h2o} and ASVD\cite{asvd}, in which the first two are token pruning methods and the last one could be regarded as a channel shrinking method\footnote{While the standard ASVD perform low-rank decomposition on all weights, here we merely decompose the $W_K,W_V$ in each layer.}. We select compression ratios of 50\% and 80\% for the experiments, with the same compression ratios for keys and values.

\subsection*{C. Ablation Study}
Without loss of generality, we perform an ablation study on LongEval with the Longchat-7b-v1.5-32k model. The window size is set to 32 and the compression ratio is evenly distributed on keys and values by default. The "Avg.Acc" column in the following tables indicates the average accuracy on the four chosen subsets of LongEval.

\subsubsection*{C.1 Effect of Initialization Methods}
\label{app:init_methods}
\begin{figure}[H]
    \centering    
    \includegraphics[width=11cm, height=6.5cm]{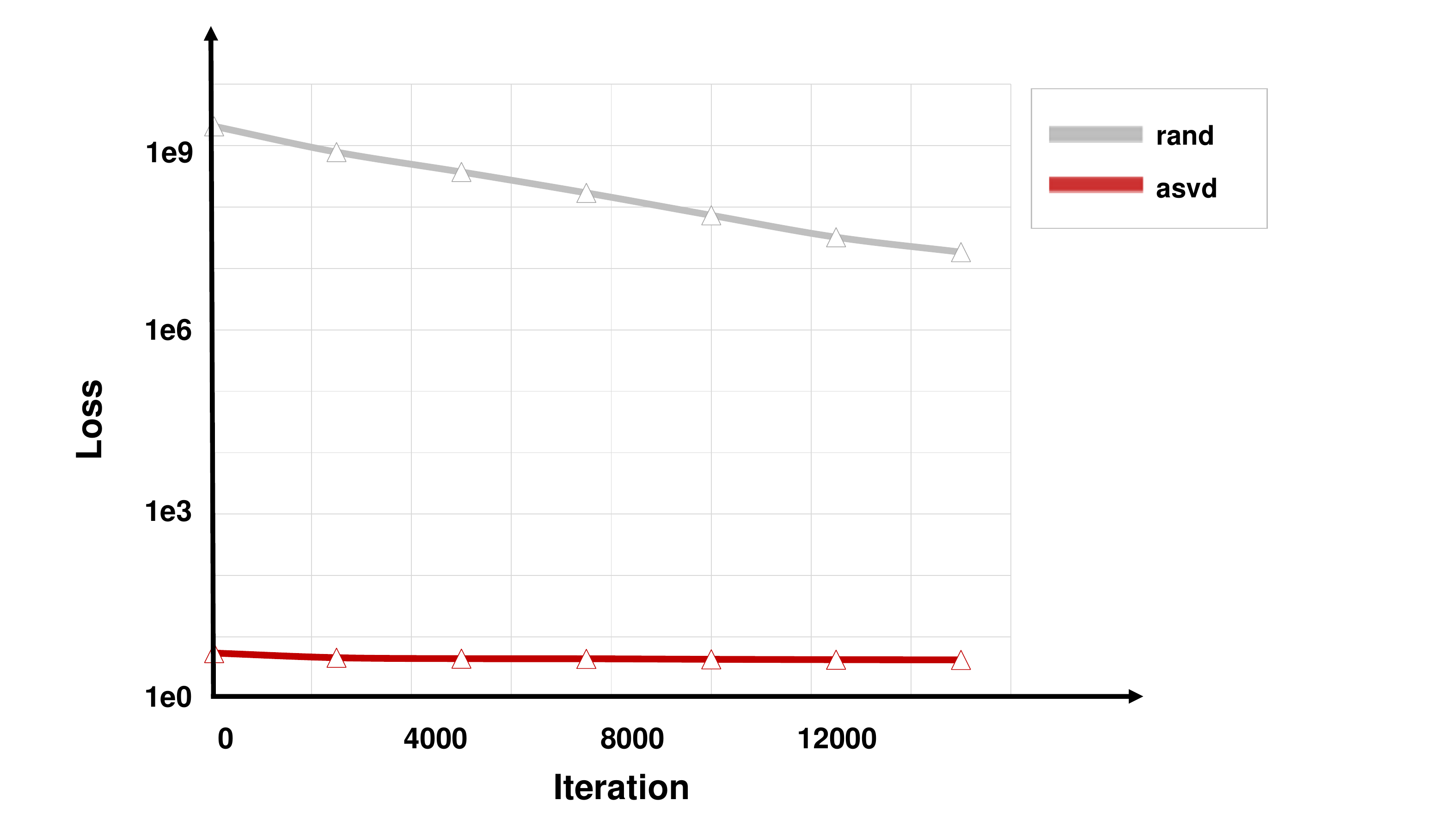}
    \caption{Loss curve with 80\% compression ratio. "asvd" means initialize with ASVD, "rand" means random initialization. We drop the curve for standard SVD initialization as it almost overlaps with the ASVD one in the figure.}
    \label{fig:loss_curve}
\end{figure}

We test three initialization methods for the low-rank decomposed matrices: 1) random initialization, 2) standard SVD initialization, and 3) ASVD initialization.
We keep their fine-tuning settings the same as mentioned in the Experimental Setups. 
The loss curves of 80\% compression are shown in Figure~\ref{fig:loss_curve}, and the evaluation results for the trained models with a bi-branch strategy are shown in Table~\ref{tab:init_method}.

\begin{table}[h]
    \centering
    \caption{Results of different initialization methods}
    \begin{tabular}{ccc}
         \toprule
         C. Ratio & Init. Method & Avg. Acc \\
         \midrule
         0\% & - & 0.99 \\
         \cmidrule(lr){1-3}
         \multirow{3}{*}{50\%} & Random & 0.00 \\
         & SVD & 0.94 \\
         & \textbf{ASVD} & \textbf{0.95} \\
         \cmidrule(lr){1-3}
         \multirow{3}{*}{60\%} & Random & 0.00 \\
         & SVD & 0.93 \\
         & \textbf{ASVD} & \textbf{0.94} \\   
        \cmidrule(lr){1-3}
         \multirow{3}{*}{70\%} & Random & 0.00 \\
         & SVD & 0.89 \\
         & \textbf{ASVD} & \textbf{0.93} \\
         \cmidrule(lr){1-3}
         \multirow{3}{*}{80\%} & Random & 0.00 \\
         & SVD & 0.87 \\
         & \textbf{ASVD} & \textbf{0.92} \\         
        \bottomrule
    \end{tabular}
    \label{tab:init_method}
\end{table}

It could be found that the loss of the random initialization method remains extremely high ($\sim$1e9) and is very hard to converge in a reasonable time, leading to the deterioration of model performance. This is quite intuitive as the information stored in the initial $W_K,W_V$ are completely destroyed and their information cannot be utilized. In contrast, the SVD-based initialization methods' loss could converge quickly from approximately 5.5 to 4.0, leading to superior model performance. \textbf{Therefore, the SVD-based initialization methods is crucial to the success of training}. Specifically, the ASVD-initialized model performs slightly better than the SVD-initialized one after training, so we choose ASVD as the default initialization method.

\subsubsection*{C.2 Effect of Window Size}
\label{app:win_size}
The window size determines how much local information could be preserved, which is of vital importance to the quality of generated content. We fix the compression ratio to 80\% and evaluate the performance of the bi-branch trained model with multiple window size settings. The results are shown in Table~\ref{tab:win_size}.

\begin{table}[h]
    \centering
    \caption{Results of different window sizes.}
    \begin{tabular}{ccc}
         \toprule
         C. Ratio & Window Size
         & Avg. Acc \\
         \midrule
         0\% & - & 0.99 \\
         \cmidrule(lr){1-3}
         \multirow{10}{*}{80\%} & 2 & 0.77 \\
         & 4 & 0.83 \\
         & 8 & 0.85 \\
         & 16 & 0.88 \\
         & \textbf{32} & \textbf{0.92} \\
         & 64 & 0.93 \\
         & 128 & 0.94 \\
         & 256 & 0.94 \\
         & 512 & 0.94 \\
         & 1024 & 0.95 \\
         & 2048 & 0.96 \\
         & 4096 & 0.96 \\
        \bottomrule
    \end{tabular}
    \label{tab:win_size}
\end{table}

The accuracy of the model shows a positive correlation with the window size, which is quite intuitive. Specifically, as the window size increases from 2 to 32, the accuracy improves relatively rapidly. However, when the window size exceeds 32, the rate of accuracy improvement notably decreases. This might indicate that a window size around 32 would be enough for local information preservation, while greater window sizes could not bring obvious improvement. \textbf{Therefore, we may conclude that the model performance is positively correlated with the window size, while the benefit would become less significant after it reaches a certain level}. Considering that an excessively large window size incurs non-negligible memory overhead, practitioners should carefully balance the trade-off between memory budget and accuracy when selecting the optimal window size for real-world applications.

\subsubsection*{C.3 Effect of Compression Ratio Allocation for KV}
\label{app:alloc_kv}
Different from the token pruning methods that have to keep or discard a certain token's keys and values simultaneously, our channel shrinking method allows for the key cache and value cache to have different compression ratios. To investigate the impact of allocating a certain compression ratio to the key cache and value cache in different proportions, we conduct experiments by fixing the total compression rate at 50\% and 75\%, respectively. We then evaluate the model's performance under various combinations of compression ratios for keys and values. The results are shown in Table~\ref{tab:kv_alloc}.

\begin{table}[h]
    \centering
    \caption{Results of different compression ratio assignments}
    \begin{tabular}{ccc}
         \toprule
         C. Ratio & KV C. Ratio
         & Avg. Acc \\
         \midrule
         0\% & - & 0.99 \\
         \midrule
         \multirow{7}{*}{50\%} & K(87.5\%) V(12.5\%) & 0.97 \\
         & \textbf{K(75.0\%) V(25.0\%)} & \textbf{0.98} \\
         & K(62.5\%) V(37.5\%) & 0.96 \\
         & K(50.0\%) V(50.0\%) & 0.95 \\
         & K(37.5\%) V(62.5\%) & 0.95 \\
         & K(25.0\%) V(75.0\%) & 0.94 \\
         & K(12.5\%) V(87.5\%) & 0.80 \\
         \midrule
         \multirow{7}{*}{75\%} & K(43.75\%) V(6.25\%) & 0.73 \\       
         & K(37.50\%) V(12.50\%) & 0.89 \\
         & \textbf{K(31.25\%) V(18.75\%)} & \textbf{0.95} \\
         & K(25.00\%) V(25.00\%) & 0.93 \\
         & K(18.75\%) V(31.25\%) & 0.88 \\
         & K(12.59\%) V(37.50\%) & 0.80 \\
         & K(6.25\%) V(43.75\%) & 0.43 \\
        \bottomrule
    \end{tabular}
    \label{tab:kv_alloc}
\end{table}

It could be found from the evaluation results that among the selected combinations, the optimal configuration consistently occurs when the compression ratio for the key cache exceeds that of the value cache, showing that \textbf{it would be better to compress the key cache more than the value cache given a certain budget, in most cases}. This potentially reveals that the sensitivity of keys towards compression is weaker than that of values, making the key cache much easier to compress.

\subsubsection*{C.4 Compatibility with Quantization}
\label{app:quant}
As the low-bit quantization methods are orthogonal with our method, we further demonstrate that quantization could be seamlessly combined with our method. Specifically, we apply KIVI~\cite{kivi} with 4-bit quantization on the compressed keys and values, using per-channel quantization for the former and per-token quantization for the latter. Both the window size and the residual size are set to 32. We separately perform the experiments with two quantization manners: PTQ (Post-Training Quantization) and QAT (Quantization-Aware Training). The results are shown in Table~\ref{tab:quant}, where the "None" rows are the referenced results from the full-precision model. 

\begin{table}[h]
    \centering
    \caption{Results of integration with quantization}
    \begin{tabular}{cccc}
         \toprule
         C. Ratio (origin) & C. Ratio (4-bit) & Q. Mode
         & Avg. Acc \\
         \midrule
         0\% & 0\% & - & 0.99 \\
         \cmidrule(lr){1-4}
         \multirow{3}{*}{50\%} & \multirow{3}{*}{87.5\%} & None & 0.95 \\
         & & PTQ & 0.00 \\
         & & \textbf{QAT} & \textbf{0.96} \\
         \cmidrule(lr){1-4}
         \multirow{3}{*}{60\%} & \multirow{3}{*}{90.0\%} & None & 0.94 \\
         & & PTQ & 0.00 \\
         & & \textbf{QAT} & \textbf{0.94} \\
         \cmidrule(lr){1-4}
         \multirow{3}{*}{70\%} & \multirow{3}{*}{92.5\%} & \textbf{None} & \textbf{0.93} \\
         & & PTQ & 0.00 \\
         & & QAT & 0.92 \\
         \cmidrule(lr){1-4}
         \multirow{3}{*}{80\%} & \multirow{3}{*}{95.0\%} & \textbf{None} & \textbf{0.92} \\
         & & PTQ & 0.00 \\
         & & QAT & 0.90 \\
        
        \bottomrule
    \end{tabular}
    \label{tab:quant}
\end{table}

According to the results in Table~\ref{tab:quant}, directly applying PTQ would completely deteriorate the model's performance, while the QAT results show minor degradation compared with their full-precision counterparts. The failure of PTQ might be a result of the significant density of the compressed representations, which are a lot more intact and difficult to directly quantize. In contrast, the QAT method includes the quantization loss during the optimization process and shows great compatibility with our channel shrinking method, where a total of 95\% compression would still keep more than 90\% of the model's long-context capability. Therefore, it could be concluded that \textbf{it would be better to compress the key cache more than the value cache given a certain budget, in most cases}.

\end{document}